\documentclass[10pt,twocolumn,letterpaper]{article}

\usepackage{cvpr}

\usepackage[dvipsnames]{xcolor}

\usepackage{makecell}
\usepackage{multirow}

\definecolor{cvprblue}{rgb}{0.21,0.49,0.74}
\usepackage[pagebackref,breaklinks,colorlinks,citecolor=cvprblue]{hyperref}

\usepackage[accsupp]{axessibility}

\def\confName{CVPR}
\def\confYear{2024}

\title{Facial Identity Anonymization via Intrinsic and Extrinsic Attention Distraction}

\author{
	Zhenzhong Kuang\textsuperscript{\rm 1}\quad
	Xiaochen Yang\textsuperscript{\rm 1}\quad
	Yingjie Shen\textsuperscript{\rm 1}\quad
	Chao Hu\textsuperscript{\rm 1}\quad
	Jun Yu\thanks{Jun Yu is the corresponding author}~~\textsuperscript{\rm 1, 2}
	\\
	\normalsize\textsuperscript{\rm 1} School of Computer Science, Hangzhou Dianzi University, Hangzhou, China\\
	\normalsize\textsuperscript{\rm 2} Department of Computer Science and Technology, Harbin Institute of Technology (Shenzhen), Shenzhen, China\\
	{\tt\small \{zzkuang, xcyang, shenyingjie, chaos, yujun\}@hdu.edu.cn}
}

\begin{document}
\maketitle
\begin{abstract}
	The unprecedented capture and application of face images raise increasing concerns on anonymization to fight against privacy disclosure. Most existing methods may suffer from the problem of excessive change of the identity-independent information or insufficient identity protection. In this paper, we present a new face anonymization approach by distracting the intrinsic and extrinsic identity attentions. On the one hand, we anonymize the identity information in the feature space by distracting the intrinsic identity attention. On the other, we anonymize the visual clues (i.e. appearance and geometry structure) by distracting the extrinsic identity attention. Our approach allows for flexible and intuitive manipulation of face appearance and geometry structure to produce diverse results, and it can also be used to instruct users to perform personalized anonymization. We conduct extensive experiments on multiple datasets and demonstrate that our approach outperforms state-of-the-art methods.
	\vspace{-0.2cm}
\end{abstract}

\section{Introduction}
\label{sec:intro}

By making full use of face images, modern AI technologies have enabled us a more convenient life \cite{News19,diniz2020face,AdaFace}. However, this may raise a wide social concern on privacy because face images are easy to capture but cannot be easily changed. Although some strict constraints (e.g. laws) were set up in the last few years \cite{CIAGAN,yang2021imagenetfaces,zhai2022a3gan}, the privacy disclosure events continue to emerge one after another.

Anonymization has attracted increasing attention, which usually has two basic requirements. The first one is to ensure identity safety by fighting against re-identification. Another is to preserve the data utility, such as image quality, face detectability, expression and user-defined attributes, which may vary under different scenarios. Besides protecting the original identity, we also take identity intrusion into consideration to reduce the risk of bringing troubles for the others. The kind of technology has multiple advantages, such as: (1) prevent unauthorized users, organizations and applications from freely collecting and using personal data; (2) help people to avoid troubles by blocking the relationship disclosure between identity and the other factors, such as location, action, event and so on; (3) maintain the data usability in various applications, like autonomous driving and remote medical system, without worrying about information leakage even if the data were attacked or misused.

Traditional methods (e.g. pixelation and blurring \cite{BlurBlock,Leverage18}), seem simple and effective for privacy protection, but may easily damage the image content and quality, resulting in poor data reusability (e.g. face may become undetectable). Recently, the generative method (e.g. GAN) show promising performances on supporting realistic face synthesis \cite{image2image,StyleGAN,gafni2019live,LDM22}, which makes it possible to improve image quality and utility preservation. On this basis, many excellent anonymization trials were conducted from different viewpoints \cite{NEO2018,ren2018learning,DeepPrivacy19,CIAGAN,zhu2020deepfakes,cao2021personalized,PINet,singh2022decouple,wen2022identitydp,LDFA23,FALCO23,Riddle23,DeepPrivacy23}. However, many of them may suffer from the problem of excessive change of the identity-independent information to ensure anonymity, or insufficient identity protection in order to preserve more data utility. The former may lead to performance drop on utility preservation, and the latter may lead to degraded protection against re-identification or identity intrusion, which would prevent the existing methods from achieving a good privacy-utility (PU) tradeoff.

\begin{figure}
	\centering
	\hfill
	\includegraphics[width=\linewidth]{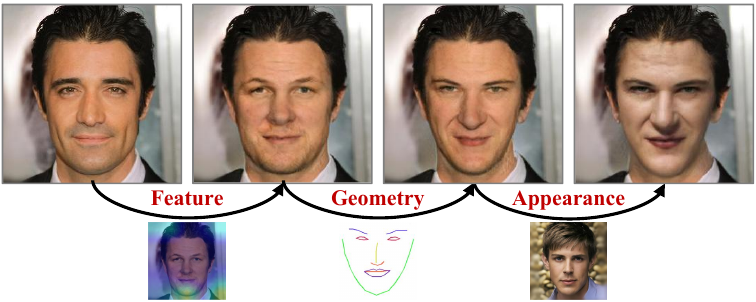}
	\vspace{-0.6cm}
	\caption{Demonstration of our approach for face anonymization.}
	\label{fig:demo}
	\vspace{-0.4cm}
\end{figure}

To address the above problem, we present a new face anonymization approach by exploiting the intrinsic and extrinsic face characteristics for identity attention distraction, where a deep generative model is employed to synthesize anonymous face images. To enable flexible control on anonymization, we divide the input data into two types, including intrinsic identity feature and extrinsic visual clues. Many works intend to embed additional PU tradeoff constrains in their model, but this may increase the difficulty of model optimization. Differently, we propose to perform data anonymization in ahead and let the deep generative model only focus on synthesizing high utility images. Since attention reflects the intrinsic characteristics of the recognition process \cite{CAM16,GradCAM,jiang2021layercam,zatorre1999auditory,wang2021dual}, we perform identity feature anonymization (IFA) by distracting the attention of the original identity to let the face recognition model make wrong prediction. Since the extrinsic face characteristics may attract human attention for re-identification, we perform visual clue anonymization (VCA) by distracting the identity attention of visual clues (i.e. visual appearance and geometry structure) that may easily lead to privacy disclosure. Figure \ref{fig:demo} briefly demonstrates the idea of our approach.

Notice that, by proper modeling, IFA can achieve a low loss of identity independent information for utility preservation. In the meanwhile, VCA can enable users more freedom in producing diverse results without significant damagement on data utility. For example, it can support fine-grained adjustment on geometry structure to support more effective anonymization, which was rarely considered previously (e.g. \cite{CIAGAN,DeepPrivacy19,k21,cao2021personalized,DIPM21,Unnoticeable,li2021identity,PROFace22,yuan2022generating,Riddle23}). During the anonymization process, our approach can enable users to easily spot what kind of changes were made by comparing input and output, which can be used to instruct users on how to perform personalized anonymization. In summary, the main contributions of this paper are as follows:
\begin{itemize}
	\item{We propose a new synthetic face anonymization approach from the viewpoint of identity attention distraction by exploiting the intrinsic and extrinsic face characteristics.}
	\item{We propose an intrinsic identity attention distraction method for IFA anonymization in the feature space.}
	\item{We propose an extrinsic identity attention distraction method for VCA anonymization in the visual space.}
	\item{We demonstrate that the proposed approach can achieve state-of-the-art performance with the help of extensive experiments on different public datasets.}
\end{itemize}

%-------------------------------------------------------------------------
\section{Related Works}
\label{sec:related_work}

\textbf{Class Activation Mapping.} Interpretability is very important for deep learning based AI systems. Visualization of CNN predictions has received wide attention to interpret deep networks \cite{mahendran2016salient,chen2019looks,jiang2021layercam}. The most relevant approach is CAM \cite{CAM16} which can highlight class-specific discriminative regions by mapping the predicted class score back to the last convolutional layer of a face classification network. In \cite{GradCAM}, CAM was generalized to gradient CAM that exhibited excellent ability in providing faithful visual explanations. In \cite{li2021identity}, CAM was used to locate and change the identity-independent regions and attributes which were utilized to anonymize face images to fool human instead of machine. But, in our study, the output of gradient CAM is used as an indicator to find and recast the identity feature to fool both human and machine, which can enable us to reduce the information loss during the anonymization process for achieving a better privacy-utility tradeoff.

\textbf{Face Synthesis.} GAN has already been used in face image synthesis by playing an adversarial game between generator and discriminator \cite{image2image,StyleGAN,DPE}. In \cite{zakharov2019few}, a landmark driven synthesis method was proposed for talking head generation. In \cite{lee2020maskgan}, MaskGAN was proposed for interactive face image manipulation. In \cite{zhu2020deepfakes}, DeepFake was used to perform face swapping to protect medical video data. In \cite{Faceshifter20}, FaceShifter was introduced to perform face swapping by focusing on identity transformation, which was further applied in \cite{cao2021personalized} and \cite{wen2022identitydp} to support face anonymization.

\textbf{Face Anonymization.} Along with the unprecedented application of face images, face anonymization becomes increasingly important and lots of interesting methods were proposed  \cite{NEO2018,DeepPrivacy19,ANet,gu2020password,CIAGAN,PrivacyNet,li2021identity,cao2021personalized,DIPM21,k21,PINet,uunet,CFANet,PROFace22,wen2022identitydp,yuan2022generating}. \cite{NEO2018,DeepPrivacy19,LDFA23} relied on inpainting to synthesize anonymous face. \cite{ANet,CIAGAN,PrivacyNet,uunet} adopted attribute editing, classifier or control vector to support face anonymization. \cite{gu2020password,cao2021personalized,uunet} studied the reversible face anonymization methods based on password or attribute vector. \cite{DIPM21,CFANet,wen2022identitydp} employed disentanglement or identity perturbation to de-identify face identity. \cite{yuan2022generating,PINet,FALCO23,Riddle23} synthesized anonymous faces in the StyleGAN latent space. \cite{PROFace22,li2021identity} only focused on fooling human eyes by preserving the original identity. Different from previous works, we present a new solution from the viewpoint of identity attention distraction.

\section{The Proposed Approach}
\label{sec:method}

In this section, we elaborate on our proposed approach. To alleviate developing complicated generative models, we propose a very simple two-step based anonymization process in Figure \ref{fig:flowchart} (a). Given a face image $x$, we rely on step1 to preprocess it by using IFA and VCA, and their outputs are used in step2 to synthesize an anonymous face $\hat{x}$, where step1 is only responsible for anonymization, step2 is only responsible for face image synthesis, and model training only happens in step2. Under this setting, we are able to reduce the difficulty of model design and optimization for processing a complicated privacy-utility tradeoff. Next, we present the details in several sections.

\subsection{Identity Feature Anonymization}
\label{sec:identity}

\begin{figure*}
	\centering
	\hfill
	\includegraphics[width=\linewidth]{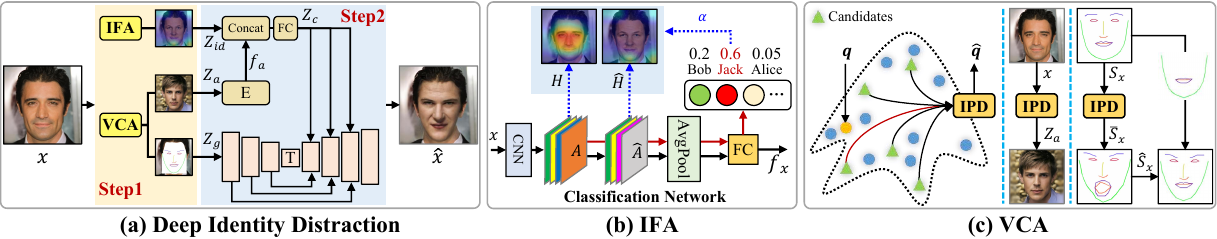}
	\vspace{-0.6cm}
	\caption{Overview of our approach: (a) the flowchart, (b) identity feature anonymization (IFA), and (c) visual clue anonymization (VCA).}
	\label{fig:flowchart}
	\vspace{-0.4cm}
\end{figure*}

In this subsection, we present how to perform identity feature anonymization in the feature space. As shown in Figure \ref{fig:flowchart} (b), a pre-trained classification network $\Psi$ (e.g. FaceNet \cite{FaceNet}) is employed for identity feature anonymization by conducting intrinsic identity attention distraction at the last convolutional layer (its output is denoted as $A$). First, the input data flow goes along the red solid lines to find the identity related feature maps in $A$ on top of the calculated CAM heatmap $H$ \cite{GradCAM}. Then, the data flow switches to the black lines after performing attention distraction on $A$, goes through $\hat{A}$, and finally output the recasted identity feature $f_{x}$ for anonymization. $\hat{A}$ is calculated from $A$ by distracting the identity attention away from $H$, where the visual results of $H$ and $\hat{H}$ in Figure \ref{fig:flowchart} (b) illustrate the function of this operation. Note that the identity of $x$ may be new for $\Psi$ and, thus, the prediction may be incorrect. But, it does not matter. We simply employ the pre-trained classes as the codebook to interpret the identity of any input faces regardless of whether these identities were trained or not. The final softmax output is used as the indicator to determine which pre-trained identities are more related to $x$.

Let $c_{i}$ denote the top $i$-th prediction result (i.e. identity class) of $\Psi$. The CAM heatmap of any given class $c_{i}$ can be calculated as the weighted combining of the forward activation maps by following \cite{GradCAM}
\vspace{-0.1cm}
\begin{equation}
	\small
	\label{equ:cam}
	H=ReLU(\alpha^{c_{i}}A)=ReLU(\sum_{j}\alpha^{c_{i}}_{j}A^{j}),
	\vspace{-0.1cm}
\end{equation}
where $\alpha^{c_{i}}_{j}$ denotes the neuron importance weight of the $j$-th activated feature map $A^{j}$, which is calculated by average-pooling the gradients flowing back
\vspace{-0.1cm}
\begin{equation}
	\small
	\label{equ:alpha}
	\alpha^{c_{i}}_{j}=\frac{1}{Z}\sum_{k}\sum_{l}\frac{\partial y^{c_{i}}}{\partial A^{j}_{kl}}.
	\vspace{-0.1cm}
\end{equation}

We analyze the importance of facial features on top of $H$. Motivated by the existing studies that some attributes are critical for identification while the others not \cite{li2021identity,GradCAM,taherkhani2018deep,diniz2020face}, we can reasonably suppose that anonymization can be formulated as a min-max optimization problem by suppressing the identity dependent attributes and preserving the identity-independent ones. We model this as an attention distraction problem by using the identity correlated CAM heatmaps and enforcing $H_{c_{i}}\rightarrow 0$. Then, we have
\vspace{-0.1cm}
\begin{equation}
	\small
	\label{equ:deact}
	\hat{A}=A+\xi, ~~~s. t. ~~~\alpha^{c_{i}}\hat{A}=0
	\vspace{-0.1cm}
\end{equation}
where $\xi$ is a modulation item. It is an invalid solution by directly using $\xi=-A$ because the resulting distracted feature map $\hat{A}$ would become meaningless for identity representation. To solve it, we introduce an assistant matrix $\Phi^{c_{i}}$ to distract the attention of the identity correlated CAM heatmaps and define $\xi=(\alpha^{c_{i}})^{T}\Phi^{c_{i}}$ by taking the importance of the activated feature map into consideration. By substituting $\xi$ back to Equ. (\ref{equ:deact}), we obtain
\vspace{-0.1cm}
\begin{equation}
	\small
	\label{equ:solve1}
	\Phi^{c_{i}}=-\frac{\alpha^{c_{i}}A}{\alpha^{c_{i}}(\alpha^{c_{i}})^{T}}.
	\vspace{-0.1cm}
\end{equation}

To enhance anonymization, we propose to jointly distract the top $K$ predictions of $\Psi$ because they may be closely related to identity representation. We redefine Equ. (\ref{equ:deact}) as
\vspace{-0.2cm}
\begin{equation}
	\small
	\label{equ:solve3}
	\hat{A}=A+\sum_{i=1}^{K}w_{i}(\alpha^{c_{i}})^{T}\Phi^{c_{i}},
	\vspace{-0.1cm}
\end{equation}
where $w_{i}$ denotes the contribution of the $i$-th item.

\subsection{Visual Clue Anonymization}
\label{sec:structure}

In this subsection, we present how to perform visual clue anonymization by extrinsic identity attention distraction using both the visual appearance and the geometry structure in the visual space. One may think of directly replace the original data $q$ with a completely (or predefined) different delegate $\hat{q}$ (e.g. from white skin to black skin) so that no one can re-identify it, but this may easily damage data utility (e.g. ethnic and expression preservation \cite{DeepPrivacy19,CIAGAN,Riddle23}). A possible better choice is to sample some $\hat{q}$ based on $q$ so that they share more identity independent information than identity dependent information in a random manner. As shown in Figure \ref{fig:flowchart} (c), we first introduce an instance-level probabilistic delegate (IPD) sampling method and then use it to anonymize the visual clues.

\textbf{IPD Sampling.}
Given data $q$, we build a candidate set $M_{q}$ by finding its top $k$-nearest neighbors in the feature space (e.g. Arcface \cite{deng2019arcface}). For each candidate $X_{i}$$\in$$M_{q}$, we rely on the simple random sampling to obtain a delegate $\hat{q}$$=$$X_{k}$$\in$$M_{q}$ according to the probability set $\{P(X_{i}),1\leq i\leq size(M_{q})|\sum_{i}P(X_{i})=1\}$, where $P(X_{i})$ is defined by following the idea of differential privacy (DP) \cite{DP07,DP08,DP09}
%\vspace{-0.1cm}
\begin{equation}
	\small
	\label{equ:prob}
	P(X_{i})=\frac{e^{[\epsilon u(q,X_{i})/(2\Delta u)]}}{\sum_{X_{j}\in M_{q}}{e^{[\epsilon u(q,X_{j})/(2\Delta u)]}}},
%	\vspace{-0.1cm}
\end{equation}
where $\epsilon$ is the privacy budget, $u$ is the utility function and $\Delta u$ is its $\ell_{1}$ sensitivity. Notice that DP received increasing attention in face anonymization since it can provide a theoretically sound privacy protection by adding random perturbation. In previous works, DP is usually used with low-level or middle-level data (e.g. pixels \cite{fanDP,croft2021obfuscation} and identity features \cite{DIPM21,PINet,wen2022identitydp}) by adding Laplace noise. Differently, we generalize it to perform instance-level data sampling to reduce the disclosure risk of the identity information.

\textbf{Visual Appearance Anonymization (VAA).} Since the visual appearance may be correlated with some useful attributes, such as ethnic and age, the significant change of it may easily damage the data utility. To solve the problem, as shown in Figure \ref{fig:flowchart} (c), we rely on IPD to sample a delegate face $Z_{a}$ by using the following utility function (i.e. $u=u_{a}$)
\vspace{-0.1cm}
\begin{equation}
	\small
	\label{equ:utility1}
	u_{a}(x,X_{i})=\frac{\max_{j}d(x,X_{j})-d(x,X_{i})}{\max_{j}d(x,X_{j})-\min_{j}d(x,X_{j})}
\vspace{0.15cm}
\end{equation}
so that $Z_{a}$ and $x$ would have a high probability to share the same set of data utility, where $d(x,X_{j})$ is the $\ell_{2}$ distance between the features of $x$ and $X_{j}$.

\textbf{Geometry Structure Anonymization (GSA).} The detected landmarks \cite{bulat2017far} are used to describe the facial geometry structure $S_{x}$ of $x$. Instead of directly modifying the landmarks (which is a complicated task to make the result look real), we prefer to perform instance-level anonymization by replacing $S_{x}$ with another realistic delegate. As shown in Figure \ref{fig:flowchart} (c), the process consists of two steps. We first rely on IPD to sample a delegate $\bar{S}_{x}$ that has the same pose as $S_{x}$ by using the following utility function (i.e. $u=u_{g}$)
%\vspace{-0.1cm}
\begin{equation}
	\small
	\label{equ:utility2}
	u_{g}(S_{x},X_{i})=\frac{d(S_{x},X_{i})-\min_{j}d(S_{x},X_{j})}{\max_{j}d(S_{x},X_{j})-\min_{j}d(S_{x},X_{j})}
%\vspace{-0.1cm}
\end{equation}
that tends to sample a more distinct geometry structure. Since this may violate the original pose and expression, we then recover them by adjusting the contour and mouth of $\bar{S}_{x}$ according to $S_{x}$ and preserving the thickness of the upper and lower lips of $\bar{S}_{x}$, resulting in $\hat{S}_{x}$. To recover the original background, we fuse $\hat{S}_{x}$ and the background $x_{b}$ of $x$ as the geometry input $Z_{g}$.
Note that $\hat{S}_{x}$ is not the same as $\bar{S}_{x}$, which would reduce the probability of identity intrusion.

\subsection{Conditional Face Synthesis}
\label{sec:synthesis}

This subsection focuses on face synthesis. As shown in the step2 of Figure \ref{fig:flowchart} (a), our generator $G$ takes appearance image $Z_{a}$, identity feature $Z_{id}$ and geometry input $Z_{g}$ as input, goes through the appearance encoder $E$ and the conditional translator $T$ to produce a realistic face image $\hat{x}$.

\textbf{Appearance Encoder $E$} is adopted to process $Z_{a}$ to obtain an appearance feature $f_{a}$. We realize it by stacking six ResBlocks \cite{resnet} and a SumPooling layer \cite{zakharov2019few}. Semantic segmentation \cite{yu2018bisenet} is used to obtain the foreground face image of $Z_{a}$ as input, when it is not available, we can approximately use the detected landmarks to realize this.

\textbf{Conditional Translator $T$} aims at translating $Z_{g}$ and the condition input $Z_{c}$ to a realistic face image $\hat{x}$. We employ a U-Net like structure to build $T$ by following \cite{UNet,image2image,zakharov2019few} via downsampling and upsampling with ResBlocks, where adaptive instance normalization (AdaIN) \cite{AdaIN} is employed to fuse the identity and appearance information encoded in $Z_{c}$ which is defined as the fusion of $Z_{id}$ and $f_{a}$ by using a concatenation (Concat) layer and a fully connected (FC) layer: $Z_{c}$=FC(Concat($Z_{id}$, $f_{a}$)). 
Note that, if the out contour were changed, $T$ would adaptively inpaint the background so that the generated face could smoothly dissolve into the original background. Users can realize personalized anonymization by manipulating $Z_{id}$, $Z_{a}$ and $Z_{g}$ under different scenarios. For example, one can simply use the facial region of $x$ to preserve the original appearance or attributes.

\subsection{Training and Optimization}
\label{sec:train}

Let $\{x, y\}$ be a set of randomly sampled face images, $x$ acts as the image to be anonymized and $y$ acts as the identity provider, we present a 1:1 alternative reconstruction and cycle swap-reconstruction strategy for network training. For the former, $\hat{x}$ is reconstructed from $x$, denoted as $Z_{id}=f_{x}$. For the latter, we change the identity from $x$ to $y$ and back to $x$ in a loop, denoted as $Z_{id}=f_{y}$. The following multi-task loss function is used to optimize our generator
\begin{equation}
	\label{equ:obj}
	\small
	L_{All}=\lambda_{1}L_{1}+\lambda_{2}L_{2}+\lambda_{3}L_{3}+\lambda_{4}L_{4}+\lambda_{5}L_{5}+\lambda_{6}L_{6},
\end{equation}
where $L_{1}$ is the adversarial loss, $L_{2}$ is the feature matching loss borrowed from \cite{wang2018high} to stabilize the training process by matching the multi-layer features of discriminator D for the input and output images, $L_{3}$ is the perceptual loss, $L_{4}$ is the appearance loss, $L_{5}$ is the identity loss, $L_{6}=\mathbb{E}[|x_{b}-g_{b}|_{1}]$ is the $\ell_{1}$ between the background images of $x$ and the generated image $g$, and $\lambda_{1}\sim\lambda_{6}$ are parameters.

\begin{figure*}
	\centering
	\includegraphics[width=\linewidth]{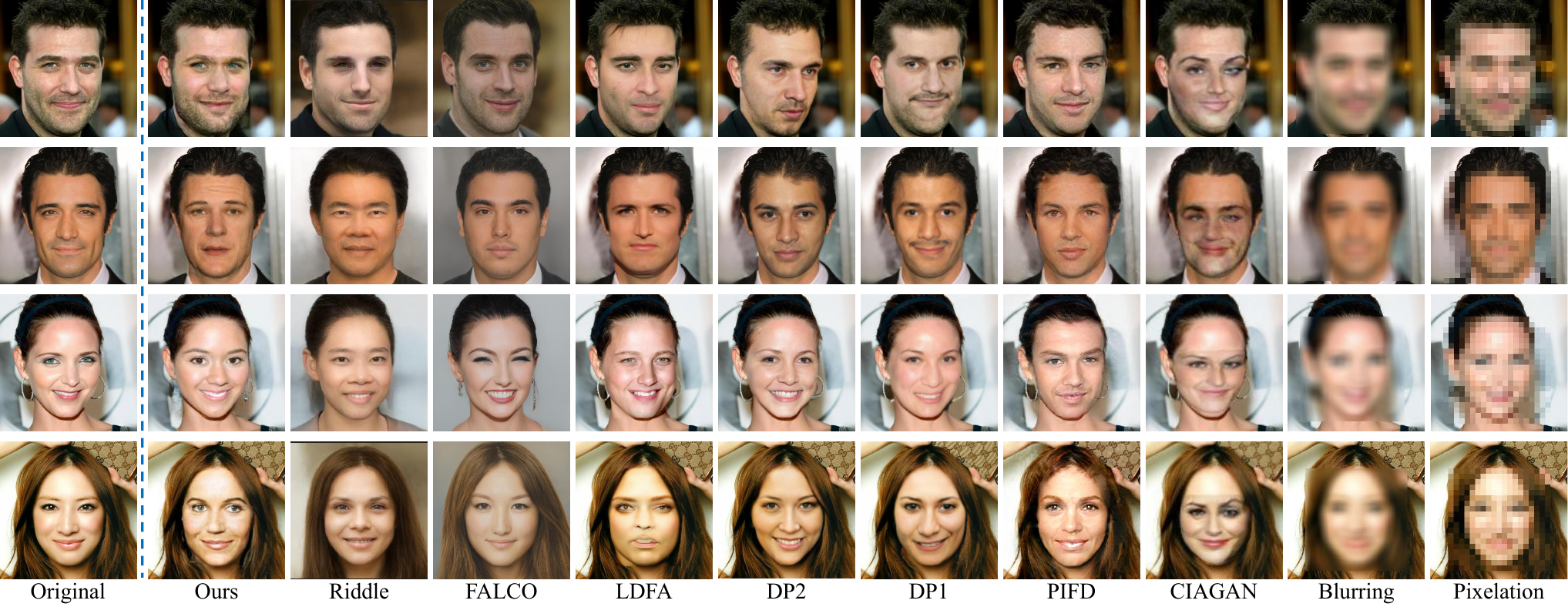}
	\vspace{-0.6cm}
	\caption{Intuitive comparison of our approach with the existing anonymization methods, where the first column presents the original faces.}
	\label{fig:compare}
	\vspace{-0.2cm}
\end{figure*}

The adversarial loss $L_{1}$ is defined as
\begin{small}
\begin{multline}
	\label{equ:loss}
	L_{1}=\left\{
	\begin{aligned}
		L_{G}(x,\hat{x})+L_{D}(\hat{x}), &~Z_{id}=f_{x}\\
		\beta_{1}(L_{G}(x,\hat{y})+L_{D}(\hat{y}))+\\L_{G}(\hat{y},\bar{x})+L_{D}(\bar{x}), &~Z_{id}=f_{y},
	\end{aligned}
	\right.
\end{multline}
\end{small}
where $L_{G}(x,\hat{x})=\mathbb{E}[\max(0,1+D(\hat{x},S_{x}))+
\max(0,1-D(x,S_{x}))]$ is used to optimize the generator G with the help of D which takes paired data (image,structure) as input, $L_{D}(\hat{x})=\mathbb{E}[-D(\hat{x},S_{x})]$ is used to optimize D, $\hat{x}=G(f_{x},Z_{a}(x),Z_{g}(x))$, $\hat{y}=G(f_{y},Z_{a}(x),Z_{g}(x))$, $\bar{x}=G(f_{x},Z_{a}(\hat{y}),Z_{g}(\hat{y}))$.
The perceptual loss $L_{3}$ is defined as
\vspace{-0.35cm}
\begin{small}
\begin{multline}
	L_{3}=\left\{
	\begin{aligned}
		&\mathbb{E}[\sum_{i}\rho_{i}(x,\hat{x})], &Z_{id}=f_{x}\\
		&\mathbb{E}[\sum\limits_{i}(\beta_{3}\rho_{i}(x,\hat{y})+\rho_{i}(\bar{x},x))], &Z_{id}=f_{y},
	\end{aligned}
	\right.
\end{multline}
\end{small}
where VGG19 and FaceNet \cite{vgg,FaceNet,johnson2016perceptual} are used to calculate $\rho_{1}$ and $\rho_{2}$.
The appearance loss $L_{4}$ is defined as
\begin{equation}
	L_{4}=\mathbb{E}[d_{a}(x,\hat{x})]+\mathbb{E}[d_{e}(x)],
\end{equation}
where $d_{a}(x,\hat{x})$ is the $\ell_{1}$ distance between $f_{a}(x)$ and $f_{a}(\hat{x})$, $d_{e}(x)$$=$$\max(0,$$cos(f_{a}(x),f_{x}))$ discourages $E$ from encoding the identity information. The identity loss is defined as
\begin{small}
\begin{multline}
	L_{5}=\left\{
	\begin{aligned}
		&\mathbb{E}[d_{f}(\hat{x},x)], &Z_{id}=f_{x}\\
		&\mathbb{E}[\beta_{2}d_{f}(\hat{x},y)+d_{f}(\bar{x},x)], &Z_{id}=f_{y},
	\end{aligned}
	\right.
	\vspace{-0.2cm}
\end{multline}
\end{small}
where $d_{f}(\hat{x},x)=|f_{\hat{x}}-f_{x}|_{2}$ is the $\ell_{2}$ distance between the identity features of $\hat{x}$ and $x$.

\section{Experiments}

In this section, we show the performance of our approach by carrying out comparative study and user study on public datasets. We also rely on ablation study to show the influences of each component of our approach.

\subsection{Settings}

\textbf{Dataset.} Three popular datasets CelebA-HQ \cite{lee2020maskgan}, VggFace2 \cite{cao2017vggface2} and LFW \cite{LFW} are used. CelebA-HQ has 30,000 high quality facial images from 6,217 persons, where 5,000 images are used as the test set. VggFace2 has 3.31 million images from 9,131 persons, where a subset of 5,000 images from 1,000 identities are used for systematical analysis and fair comparison. LFW has 13,233 face images from 5,749 individuals, where 1,680 identities have more than one image and we use a subset of 5,000 images from them.

\textbf{Implementation Details.} For pre-processing, we employ \cite{bulat2017far} to detect facial landmarks and BiSeNet \cite{yu2018bisenet} to perform semantic segmentation. We rely on \cite{image2image} and \cite{zakharov2019few} to build G and D by stacking ResBlocks. We train our network to generate 256$\times$256 images by using the Adam optimizer. We set $\lambda_{1}$$\sim$$\lambda_{4}$=1, $\lambda_{5}$=$\lambda_{6}$=$\beta_{2}$=2, $\beta_{1}$=0.6, and $\beta_{3}$=0.8. Our approach is trained on CelebA-HQ and evaluated on all.

\textbf{Evaluation Measures.} Our approach is evaluated from the perspectives of privacy protection (anonymization and identity intrusion) and data reusability. For anonymization, we calculate re-identification (ReID) rate (in percentage $\%$). For identity intrusion, we calculate identity swapping (IDS) rate ($\%$). The pre-trained FaceNet \cite{FaceNet} are ArcFace \cite{deng2019arcface} are used for face verification. The cosine similarity is used for ArcFace with two thresholds $0.30$ and $0.35$, and the $\ell_{2}$ distance is used for FaceNet with three thresholds $0.9$, $1.0$ and $1.1$. Face alignment \cite{bulat2017far} is used to evaluate face detection rate ($\%$). LPIPS and SSIM \cite{cao2021personalized} are used to evaluate image quality. Pre-trained classifiers \cite{resnet} are used to evaluate attribute preservation ($\%$), including expression, ethnic, gender, age and makeup.

\subsection{Main Results}

We mainly compare our approach with the following representative and state-of-the-art (SOTA) methods: CIAGAN \cite{CIAGAN}, PIFD \cite{cao2021personalized}, DeepPrivacy (DP1) \cite{DeepPrivacy19}, DeepPrivacy2 (DP2) \cite{DeepPrivacy23}, LDFA \cite{LDFA23}, FALCO \cite{FALCO23} and Riddle \cite{Riddle23}.

\textbf{Qualitative Results.} As shown in Figure \ref{fig:compare}, the success of Blurring and Pixlation can be contributed to the destruction of image content, which would tell the observers that the data is under protection. In contrast, the generative methods not only show excellent anonymization performance but also show higher probability of making the results imperceptible to observers. DP1 and DP2 may fail to retain some facial attributes, like expression. CIAGAN and LDFA may generate distorted faces. PIFD may bring some artifacts. FALCO and Riddle may lose the facial details. Compared with the other methods, our results can not only look realistic but also preserve more original attributes.

\begin{table}
	\centering
	\caption{Results on privacy protection: (ReID, IDS). Lower ReID and IDS rates indicates better performance.}
	\vspace{-0.3cm}
	\label{tab:compare_reid}
	\footnotesize
	\setlength{\tabcolsep}{0.8mm}{
		\begin{tabular}{lccccc}
			\toprule
			\multirow{2}{*}{Method}&\multicolumn{2}{c}{ArcFace$\downarrow$}&\multicolumn{3}{c}{FaceNet$\downarrow$}\\
			\cmidrule(lr){2-3}\cmidrule(lr){4-6}
			&$cos>$0.30&$cos>$0.35&$\ell_{2}<$0.9&$\ell_{2}<$1.0&$\ell_{2}<$1.1\\
			\midrule
			CIAGAN
			&(1.7, 93.3)&(0.5, 51.4)&(0.5, 87.0)&(3.2, 99.9)& (11.3, \textbf{100})\\
			PIFD
			&(\textbf{0.0}, 79.3)&(\textbf{0.0}, \textbf{19.1})& (\textbf{0.0}, 58.6)&(\textbf{0.0}, 99.7)&(\textbf{0.0}, \textbf{100})\\
			DP1
			&(0.9, \textbf{70.9})&(0.2, 19.4)&(1.2, 73.5) &(4.8, 99.9)&(15.0, \textbf{100})\\
			DP2&(0.5, 78.5)&(0.1, 26.4)&(0.6, 67.4)&(3.7, 99.8)& (7.4, \textbf{100})\\
			LDFA&(4.9, 81.3)&(2.1, 33.6)&(4.3, 74.6)&(15.9, 99.8)& (24.8, \textbf{100})\\
			FALCO&(6.7, 81.8)&(2.7, 37.0)&(1.4, 83.7)&(1.7, 99.9)& (5.4, \textbf{100})\\
			Riddle
			&(\textbf{0.0}, 72.9)&(\textbf{0.0}, 26.2)&(0.6, 77.7)&(1.2, 99.8)& (2.3, \textbf{100})\\
			Ours
			&(\textbf{0.0}, 78.1)&(\textbf{0.0}, 19.6)&(0.1, \textbf{34.9})&(0.3, \textbf{98.1})&(1.0, \textbf{100})\\
			\bottomrule
		\end{tabular}
	}
\end{table}

\begin{table}[!htbp]
	\centering
	\caption{The ReID and IDS results of using the Adaface model.}
	\vspace{-0.3cm}
	\label{tab:compare_adaface}
	\footnotesize
	\setlength{\tabcolsep}{1.1mm}{
		\begin{tabular}{ccccccccc}
			\toprule
			&CIAGAN&PIFD&DP1&DP2&FALCO&LDFA&Riddle&Ours\\
			\midrule
			ReID&0.7&0.3&0.8&0.6&8.0&18.3&0.4&\textbf{0.2}\\
			IDS&9.0&7.6&6.3&\textbf{5.0}&17.0&12.7&7.2&5.7\\
			\bottomrule
		\end{tabular}
	}
\end{table}

\begin{table}
	\centering
	\caption{Results on preserving facial attributes and image quality.}
	\vspace{-0.3cm}
	\label{tab:compare_utility}
	\footnotesize
	\setlength{\tabcolsep}{0.85mm}{
		\begin{tabular}{lccccccc}
			\toprule
			\multirow{2}{*}{Method}&\multicolumn{5}{c}{Attribute$\uparrow$}&\multirow{2}{*}{LPIPS$\downarrow$}&\multirow{2}{*}{SSIM$\uparrow$}\\ 
			\cmidrule(lr){2-6}
			&Express.&Ethnic&Gender&Age&Makeup&&\\
			\midrule
			CIAGAN
			&78.7&46.7&80.9&81.3&74.7&0.558&0.358\\
			PIFD
			&82.3&48.5&84.4&82.9& 63.7&0.124& 0.771\\
			DP1&54.8&52.0&84.7&84.4&66.8&0.192&0.785\\
			DP2&59.1&\textbf{52.6}&84.3&85.3&78.9&0.127&0.779\\
			LDFA&76.1&48.8&77.9&78.8&74.2&0.124&0.733\\
			FALCO&82.6&51.8&84.8&\textbf{86.3}&77.6&0.307&0.475\\
			Riddle&77.9&41.8&81.0&84.4&68.9&0.300&0.530\\
			Ours
			&\textbf{84.2}&51.5&\textbf{85.1}&83.3&\textbf{80.1}&\textbf{0.120}&\textbf{0.799}\\
			\bottomrule
		\end{tabular}
	}
\end{table}

\textbf{Quantitative Results.} Since the image content of Pixelation and Blurring is significantly destroyed, we only compare with the generative methods in Table \ref{tab:compare_reid}. Ours, Riddle \cite{Riddle23} and PIFD \cite{cao2021personalized} outperform the other methods on ReID across different face recognition backbones, especially when larger thresholds are used, where DP2 exhibits competitive results. Besides, it is very important to see if there happens identity intrusion after anonymization. CIAGAN suffers from the highest IDS rates. Compared with the best performed PIFD and DP1, our approach obtains competitive results (e.g. $58.6\%$ vs. $34.9\%$ for PIFD and Ours), which can be contributed to identity attention distraction. Also, we test another recent AdaFace model \cite{AdaFace} for face verification. According to Table \ref{tab:compare_adaface}, it is easy to observe that our approach exhibits similar behavior to that in Table \ref{tab:compare_reid}.

\textbf{Utility Preservation.} We compare the data utility of different methods in Table \ref{tab:compare_utility}. DP1 and DP2 performs poor on expression. PIFD, DP1 and Riddle perform poor on makeup. All these methods perform poor on ethnic. Compared with the other methods, our approach achieves a much better balance on all items and performs well on preserving expression and makeup. We also find that, as with SOTA methods, our approach can also achieve $100\%$ face detection rate, which also reveals its high data utility.

\begin{figure}
	\centering
	\includegraphics[width=\linewidth]{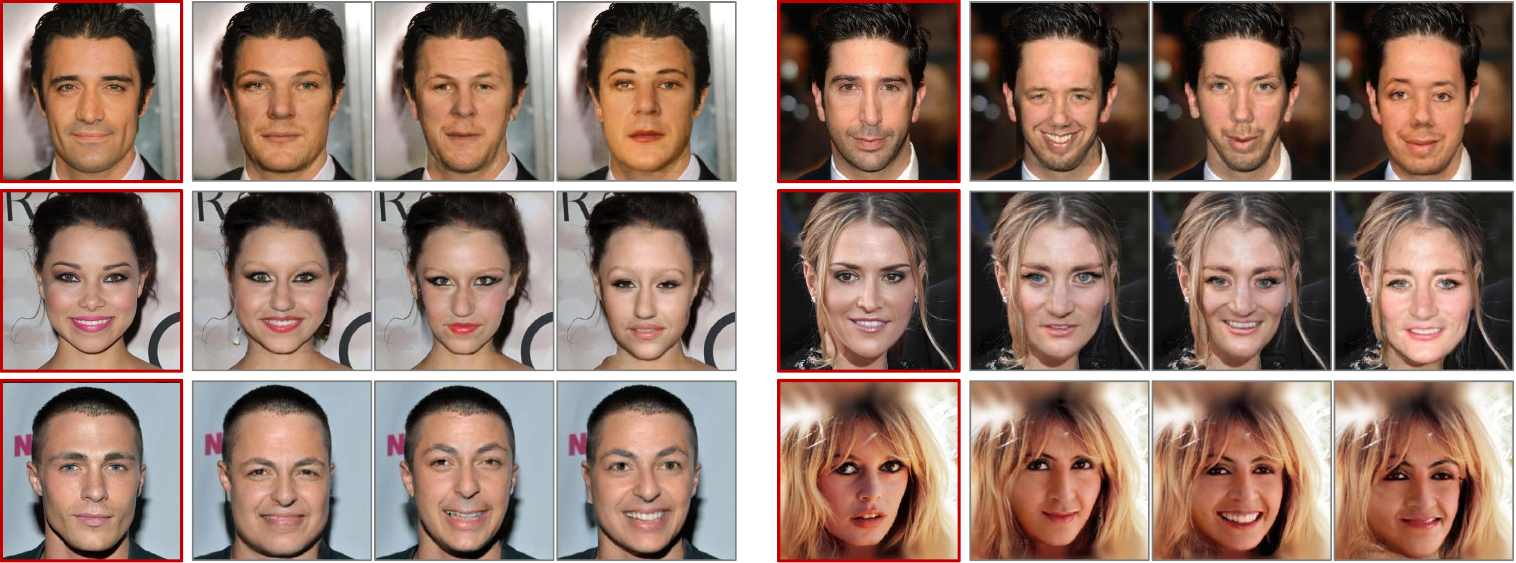}
	\vspace{-0.6cm}
	\caption{Demonstration of our diverse anonymous results.}
	\label{fig:diverse}
	\vspace{-0.2cm}
\end{figure}

\begin{figure}
	\centering
	\includegraphics[width=\linewidth]{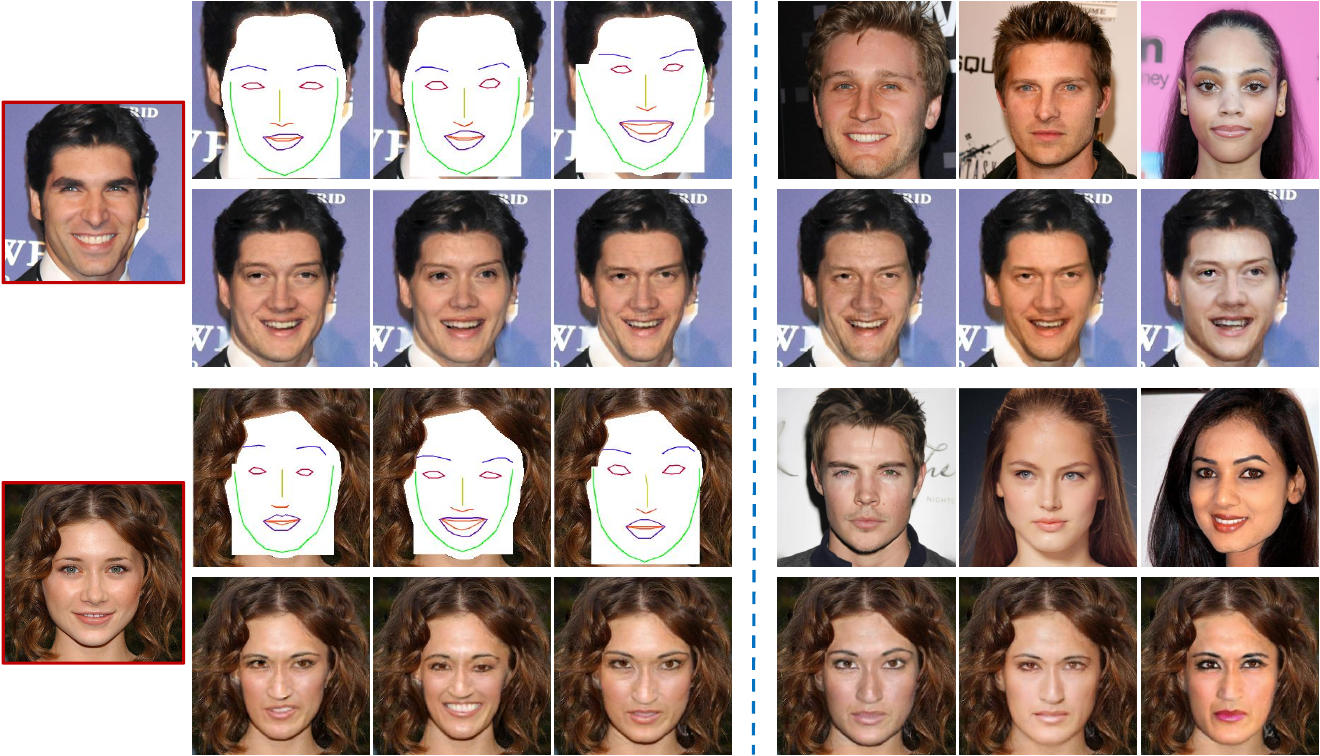}
	\vspace{-0.6cm}
	\caption{Demonstration of our results on controlling the geometry structures and visual appearances. Each of the top row shows the employed geometry structures and visual appearance images.}
	\label{fig:userdemand}
	\vspace{-0.3cm}
\end{figure}

\textbf{Diversity and Controllability.} As shown in the first two rows of Figure \ref{fig:diverse}, our approach can produce diverse results that look different from each other. Besides, we have tried to add a similar distraction item to Equ(\ref{equ:solve3}) to test the diversity of $\hat{A}$ by using the bottom $j$-th prediction ($1\leq j\leq 3$). According to the last row of Figure \ref{fig:diverse}, although we can produce diverse results, the facial expression has changed from non-smile to smile, which indicates that adding such diversity may affect the data utility. Our approach can also support flexible anonymization according to user requirements and practical applications. For example, in Figure \ref{fig:userdemand}, we can produce different anonymous faces by controlling the geometry and visual appearance inputs, where the influences of changing the geometry structure is more significant.

\textbf{Analysis.} According to the above results, it is obvious that achieving high anonymization performance is relatively easy but it is somewhat difficult to (a) prevent identity intrusion and (b) preserve data utility. The reason why (a) is hard lies in that it is uneasy to ensure the non-existence of a synthesized identity in reality. The reason why (b) is hard lies in that identity is closely related to some critical facial attributes and the change of identity would inevitably lead to some variations on facial attributes. In Figure \ref{fig:compare}, \ref{fig:diverse} and \ref{fig:userdemand}, we can intuitively observe the attribute changes on the faces (e.g. eye and nose) and they may vary for different persons. Most existing methods pay much less attention on this and anonymization is usually achieved at the cost of damaging too much useful information. Although our approach can not perform the best all the time, it has achieved a better privacy-utility tradeoff. This can be mainly contributed to our anonymization strategy of minimizing the changes on identity independent attributes. But, our approach still suffers from the drawbacks of well preserving some facial attributes, such as ethnic and eye gaze direction, which are left for the follow up works.

\begin{figure}
	\centering
	\includegraphics[width=\linewidth]{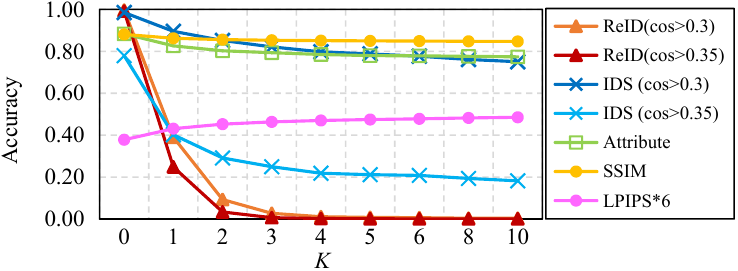}
	\vspace{-0.7cm}
	\caption{The influential curves of identity attention distraction in the feature space by increasing $K$ from $1$ to $10$. }
	\label{fig:dact}
	\vspace{-0.2cm}
\end{figure}

\subsection{Ablation Study}

In Figure \ref{fig:dact}, we plot the influential curves for the top $K$ predictions of the face classification network. With top 1 distraction, our approach can remove over $75\%$ of identity information. When increasing $K$, the ReID and IDS rates keep decreasing and the trend would slow down when $K\geq 2$. The utility preservation performance would decrease slightly along with $K$ (see Attribute and LPIPS). Most of the curves would become almost flat after $K\geq 4$. Thus, we generally recommend to set $K$ vary from $2$ to $4$ to reduce the computational costs and the loss on data utility.

\begin{table}
	\centering
	\caption{Ablation study results by removing the key components.}\vspace{-0.3cm}
	\label{tab:ablation}
	\footnotesize
	\setlength{\tabcolsep}{1.0mm}{
		\begin{tabular}{lccccc}
			\toprule
			Method&ArcFace$\downarrow$&FaceNet$\downarrow$&Attribute$\uparrow$&LPIPS$\downarrow$&SSIM$\uparrow$\\
			\midrule
			w./o. IFA
			&(45.5, 54.0)&(67.6, 85.8)&\textbf{88.0}&0.085&\textbf{0.860}\\
			w./o. VAA
			&(0.2, 20.1)&(0.2, 36.1)&77.2&\textbf{0.120}&0.800\\
			w./o. GSA
			&(0.2, 22.7)&(\textbf{0.1}, 36.6)&78.2&0.121&0.810\\
			Full Model
			&(\textbf{0.1}, \textbf{19.6})&(\textbf{0.1}, \textbf{34.9})&76.8&\textbf{0.120}&0.799\\
			\bottomrule
		\end{tabular}
	}
	\vspace{-0.1cm}
\end{table}

\begin{figure}
	\centering
	\includegraphics[width=\linewidth]{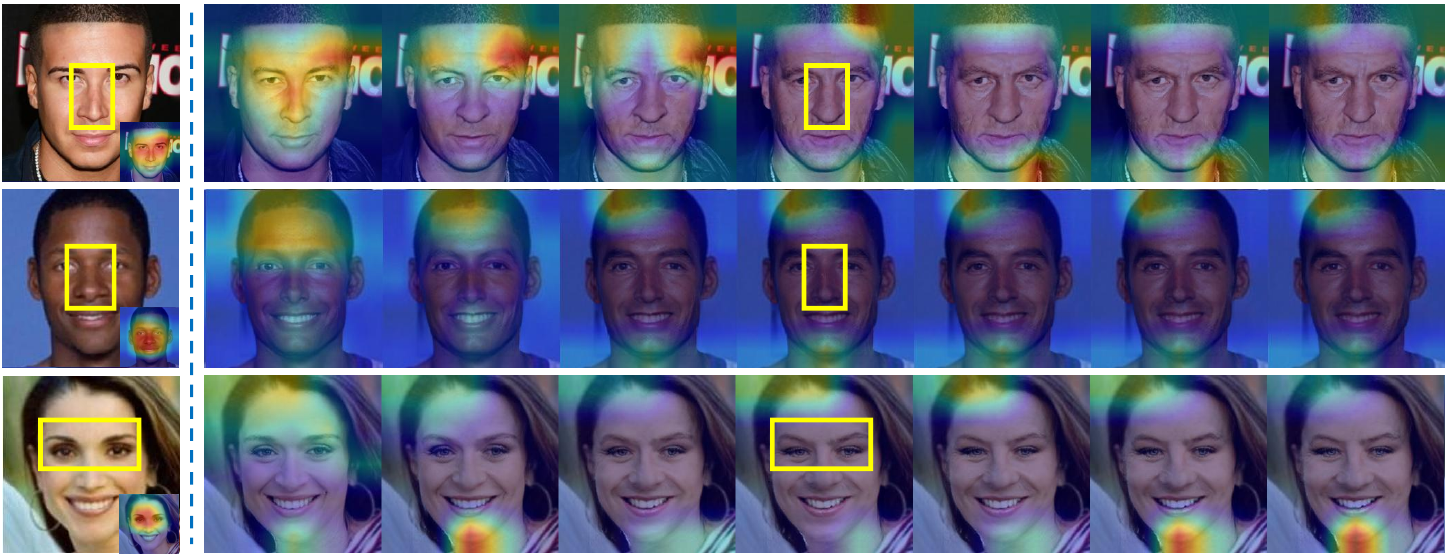}
	\vspace{-0.6cm}
	\caption{Visualization of the distracted CAM heatmaps results by increasing $K$ from $1$ to $7$ in Equation (\ref{equ:solve3}).}
	\label{fig:IDAimage}
	\vspace{-0.2cm}
\end{figure}

In Figure \ref{fig:IDAimage}, we qualitatively present some visual comparison results. With larger $K$, the CAM heatmap would focus much farther from the facial parts, especially the eyes. By comparing the face images before and after feature distraction, one can find some significant changes on the facial parts (e.g. the nose may vary from small to big or vice versa) but they may vary for different persons. The results also show that the joint distraction in Equation (\ref{equ:solve3}) would push some critical facial features or attributes heading to the opposite direction to realize identity anonymization, but this may also lead to some other unexpected changes, such as the age of the first person in Figure \ref{fig:IDAimage}. This negative effect may come from the significant change of the identity related information in $A$. These observations show that some facial parts or attributes are critical because they are correlated to identity representation, and the change of them may more or less lead to the performance drop on utility preservation.

\begin{figure}[t]
	\centering
	\includegraphics[width=\linewidth]{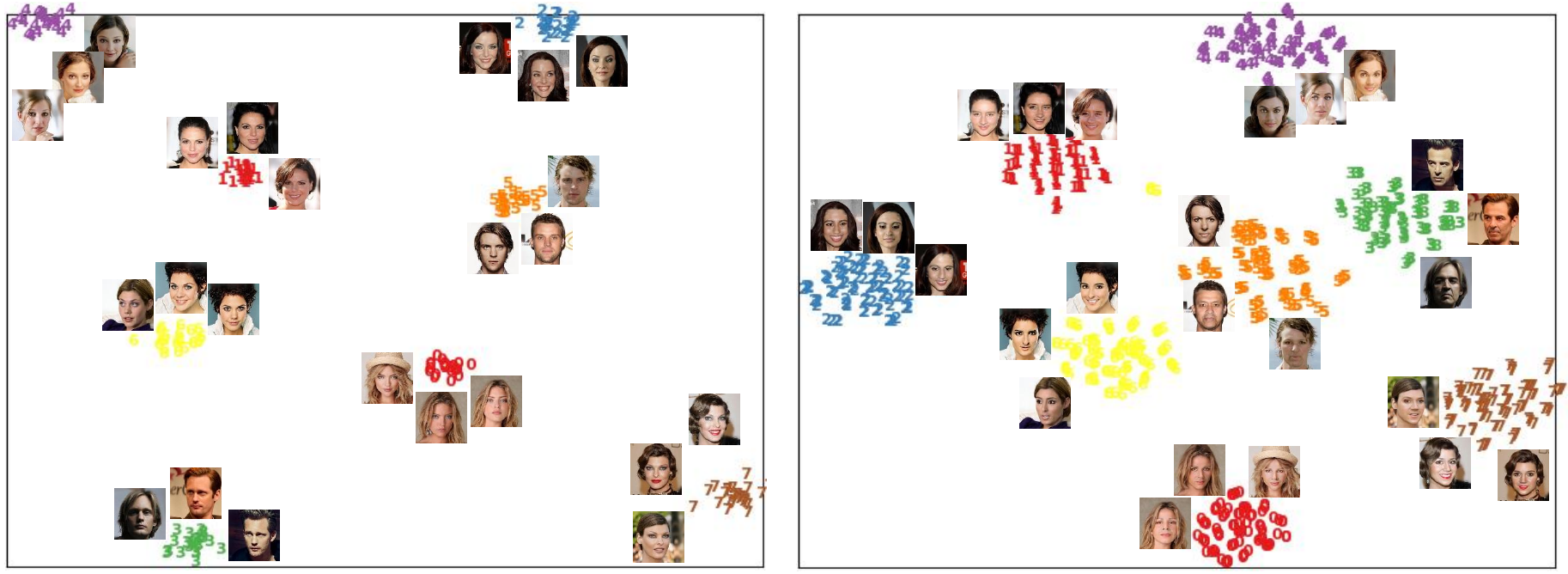}
	\vspace{-0.6cm}
	\caption{Visualization of our results before (left) and after (right) applying IFA. The colored numbers denote different persons.}
	\label{fig:cluster}
	\vspace{-0.2cm}
\end{figure}

In Figure \ref{fig:cluster}, we compare the identity features before and after IFA by using the classical T-SNE embedding \cite{tsne}. The intra-class differences would increase after identity feature distraction, but they still exhibit clustering characteristics regardless of some outliers, which would favor utility preservation. Note that our results can preserve the personalized facial attributes according to the status of each image. For example, as for the $7$-th person, our results can still retain the makeup of each instance.

In Table \ref{tab:ablation}, we present the ablation study results by removing the key components. w./o. IFA means removing the identity input and only using $Z_{a}$ as the conditional input. w./o. VAA means feeding the original face appearance to $Z_{a}$. w./o. GSA means feeding the original geometry structure to $Z_{g}$. IFA can significantly help to reduce the ReID and IDS rates, but it suffers from performance drops on attribute preservation and image quality. GSA and VAA can further help to improve the ReID and IDS performance, but may lead to some drops on utility preservation. We have also verified the expression recovery in GSA by removing it and observed a significant performance drop.%, almost $20\%$.

\begin{table}
	\centering
	\caption{Quantitative performance study on IFA and GSA.}
	\vspace{-0.3cm}
	\label{tab:GSA}
	\footnotesize
	\setlength{\tabcolsep}{1.0mm}{
		\begin{tabular}{lccccc}
			\toprule
			Method&ArcFace$\downarrow$&FaceNet$\downarrow$&Attribute$\uparrow$&LPIPS$\downarrow$&SSIM$\uparrow$\\
			\midrule%
			IFA-Rand&(0.5, 22.8)&(1.1, \textbf{36.1})&\textbf{79.7}&\textbf{0.111}&\textbf{0.821}\\
			IFA-KFN&(\textbf{0.2}, 23.1)&(\textbf{0.1}, 37.3)&79.2&\textbf{0.111}&\textbf{0.821}\\
			IFA-Ours&(\textbf{0.2}, \textbf{21.8})&(\textbf{0.1}, 36.3)&79.1&0.112&0.819\\
			\hline
			GSA-Rand
			&(45.5, \textbf{89.7})&(47.6, \textbf{79.1})&79.9&0.117&0.807\\
			GSA-KFN&(\textbf{30.7}, 89.9)&(56.7, 81.0)&80.2&0.117&0.808\\
			GSA-Ours&(43.9, 90.7)&(\textbf{38.5}, 83.2)&\textbf{86.6}&\textbf{0.083}&\textbf{0.853}\\
			\bottomrule
		\end{tabular}
	}
	\vspace{-0.3cm}
\end{table}

In Table \ref{tab:GSA}, we study the performance of IFA and GSA by using different strategies. IFA-Rand, IFA-KFN and IFA-Ours denote using random delegate, $k$-th farthest neighbor and our method to anonymize the identity feature, respectively. GSA-Rand, GSA-KFN and GSA-Ours denote using random delegate, $k$-farthest neighbor and our IPD method to anonymize the geometry structure, respectively. It is obvious that IFA-Ours and GSA-Ours have achieved a better balance between privacy protection and utility preservation.

In Equ.(\ref{equ:utility1}) and Equ.(\ref{equ:utility2}), $u_{a}$ and $u_{g}$ are determined by jointly considering anonymity and data utility. According to Table \ref{tab:uag}, $u_{a}$ can help to improve the data utility and $u_{g}$ can help to improve the protection ability.

\begin{table}[t]
	\centering
	\caption{Quantitative performance study on $u_{a}$ and $u_{g}$.}\vspace{-0.3cm}
	\label{tab:uag}
	\footnotesize
	\setlength{\tabcolsep}{0.7mm}{
		\begin{tabular}{l|ccc|ccc}
			\toprule
			Methods&ArcFace$\downarrow$&FaceNet$\downarrow$&AdaFace$\downarrow$&Attribute$\uparrow$&LPIPS$\downarrow$&SSIM$\uparrow$\\
			\midrule
			w./o. $u_{a}$
			&(\textbf{0.0}, \textbf{19.1})&(\textbf{0.0}, 36.2)&(\textbf{0.1}, 6.6)&72.3&0.160&0.750\\
			w./o. $u_{g}$
			&(0.1, 75.1)&(0.1, 36.4)&(0.7, \textbf{5.5})&\textbf{77.7}&0.128&0.784\\
			Ours
			&(\textbf{0.0}, 19.6)&(0.1, \textbf{34.9})&(0.2, 5.7)&76.8&\textbf{0.120}&\textbf{0.799}\\
			\bottomrule
		\end{tabular}
	}
	\vspace{-0.1cm}
\end{table}

\begin{figure}
	\centering
	\includegraphics[width=\linewidth]{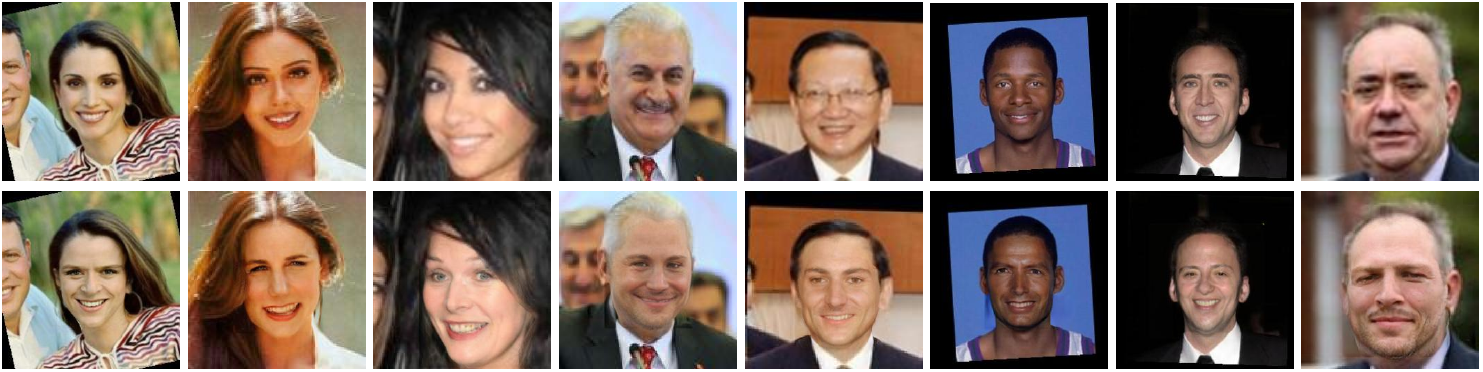}
	\vspace{-0.6cm}
	\caption{Our results (bottom) on the Vggface2 and LFW datasets.}
	\label{fig:more}
%	\vspace{-0.2cm}
\end{figure}

\begin{table}
	\begin{center}
		\caption{Results on Vggface2 (top) and LFW (bottom) datasets.}
		\vspace{-0.3cm}
		\label{tab:more}
		\small
		\setlength{\tabcolsep}{0.5mm}{
			\begin{tabular}{lccccc}
				\toprule
				&ArcFace$\downarrow$&FaceNet$\downarrow$&Attribute$\uparrow$&LPIPS$\downarrow$&SSIM$\uparrow$\\
				\midrule
				CIAGAN &(0.4, 21.6)&(1.7, 53.2)& 66.0&0.443& 0.419 \\
				PIFD&(\textbf{0.0}, \textbf{10.2})&(0.2, 38.3)&68.0&0.520&0.363\\
				DP1 &(0.4, 39.1)&(0.7,76.4)&74.5&\textbf{0.082}&\textbf{0.891}\\
				DP2&(\textbf{0.0}, 14.1)&(0.9, 58.7)&62.2&0.479&0.520\\
				Riddle&(\textbf{0.0}, 24.3)&(\textbf{0.0}, 77.9)&58.8&0.455&0.502\\
				Ours&(\textbf{0.0}, 10.7)&(0.1, \textbf{27.8})&\textbf{75.6}&0.200&0.838\\
				\hline
				CIAGAN &(0.6, \textbf{3.9})&(1.5, 27.3)&72.2&0.458& 0.397 \\
				PIFD &(\textbf{0.0}, 4.8)&(0.3, \textbf{8.0})& 66.8&0.591 & 0.314 \\
				DP1 &(0.4, 8.7)&(10.3, 61.6)&79.2&0.077&\textbf{0.895}\\
				DP2&(0.5, 7.3)&(8.3, 56.9)&79.6&\textbf{0.071}&0.804\\
				Riddle &(\textbf{0.0}, 8.5)&(\textbf{0.1}, 43.9)&61.7&0.460& 0.428 \\
				Ours &(\textbf{0.0}, 4.7)&(0.7, 16.1)&\textbf{80.9}&0.074&\textbf{0.895}\\
				\bottomrule
			\end{tabular}
		}
	\end{center}
	\vspace{-0.5cm}
\end{table}

\subsection{Results on Vggface2 and LFW}
%\vspace{-0.1cm}
We show the generalization ability of our approach on Vggface2 and LFW datasets. According to Table \ref{tab:more}, most methods show quite high privacy protection ability (e.g. $0.0\%$ ReID rate). Our approach outperforms the contrast methods on attribute, LPIPS and SSIM, which indicates that our approach has achieved a better privacy-utility tradeoff. According to Figure \ref{fig:more}, one can find that the anonymized faces on both datasets look realistic and different from their original versions. These results are in consistent with the results reported in the previous subsections, which have again helped to verify the performance of our approach.

\subsection{User Study}
\begin{figure}
	\centering
	\includegraphics[width=\linewidth]{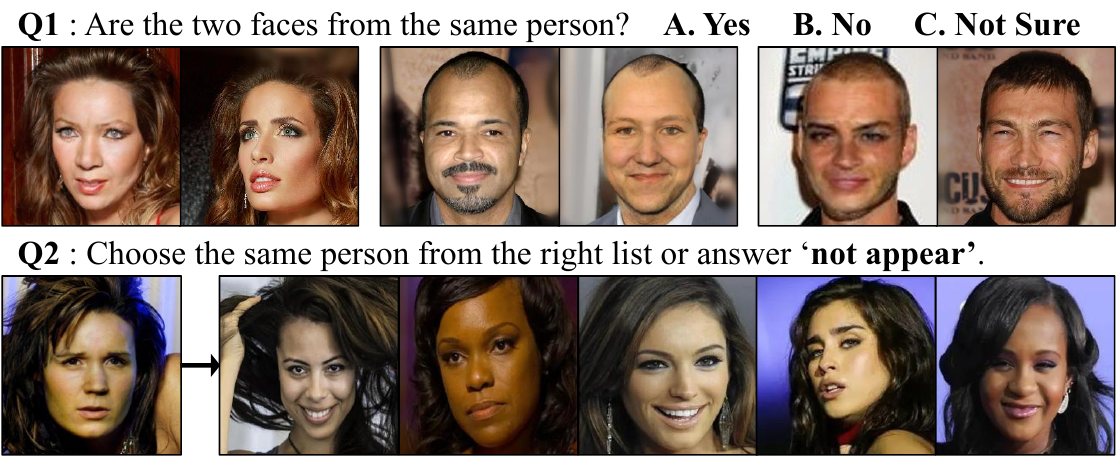}
	\vspace{-0.6cm}
	\caption{The employed questionnaires for user study.}
	\label{fig:HPS_Question}
	%\vspace{-0.1cm}
\end{figure}

We conduct a simple user study to verify the performance of our approach from the perspectives of human. As shown in Figure \ref{fig:HPS_Question}, we asked around 30 participants to answer two kinds of questionnaires: for Q1, each anonymized face is paired with another image of the original face; for Q2, the top 5 retrieved results are presented. For each of the contrast methods, we randomly assign each participant: (1) $20$ Q1 to calculate the ReID rate of choosing B and C; (2) $20$ Q2 from dataset retrieval to calculate the IDS rate of choosing 'not appear'; (3) $20$ Q2 from Google retrieval to calculate the ReID rate of choosing 'not appear'.

\begin{figure}
	\centering
	\includegraphics[width=\linewidth]{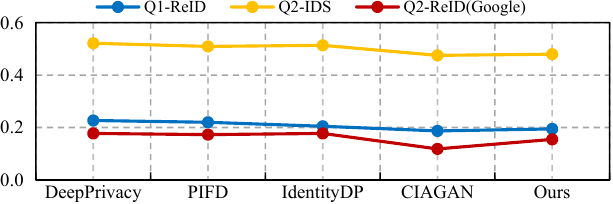}
	\vspace{-0.6cm}
	\caption{The human evaluation results on ReID and IDS.}
	\label{fig:HPS-Result}
	\vspace{-0.2cm}
\end{figure}

As shown in Figure \ref{fig:HPS-Result}, one can observe that: (1) our ReID and IDS results closely follow CIAGAN, which work better than the other methods; (2) the IDS rate of all the generative methods are high (over $45.0\%$), which may easily lead to identity intrusion. Since the image distortion would prevent the observers from correct recognition, it is easy for CIAGAN to achieve the best performance. This study show consistent results with that of machine recognizers.

\section{Conclusion}

In this paper, we present a distinct face anonymization approach from another viewpoint based on identity attention distraction. On top of ablation study, we have showed how and why our approach works. By performing comparative study and user study, we have validated our approach for improving the performance of privacy-utility tradeoff. Our approach allows for flexible manipulation of the facial appearance and geometry structure for more diverse anonymization and it has also demonstrated the generalizability in the other datasets. Future work includes exploring the correspondences between the convolutional feature maps and facial attributes for more effective anonymization, and exploring how to retain some more complex signal (e.g. psychological and physiological) hidden in visual data.

\section*{Acknowledgement}
This work was supported in part by National Natural Science Foundation of China (Grant No. 62372147, 62125201, U21B2040), Zhejiang Provincial Natural Science Foundation of China (Grant No. LY22F020028).
{
	\small
	\bibliographystyle{ieeenat_fullname}
	\bibliography{main}
}

\end{document}